\documentclass[runningheads]{llncs}
\usepackage[T1]{fontenc}
\usepackage{cite}
\usepackage{amsmath,amssymb,amsfonts}

\usepackage{amsthm}
\usepackage{algorithm}
\usepackage{algorithmic}
\usepackage{graphicx}
\usepackage{textcomp}
\usepackage{xcolor}
\usepackage{booktabs}
\usepackage[table]{xcolor}
\usepackage[caption=false,font=footnotesize]{subfig}
\usepackage{url}

\DeclareMathAlphabet{\mathbbb}{U}{bbold}{m}{n}  

\DeclareMathOperator*{\tr}{tr}

\newcommand\tsup[2][2]{%
	\def\useanchorwidth{T}%
	\ifnum#1>1%
	\stackon[-1.3ex]{\tsup[\numexpr#1-1\relax]{#2}}{\mathchar"307E\kern-.5pt}%
	\else%
	\stackon[-1ex]{#2}{\mathchar"307E\kern-.5pt}%
	\fi%
}

\newcommand{\KL}{D}

\definecolor{Gray}{gray}{0.95}

\begin{document}

\title{OSDTW: Optimal Shared Depth and Task Weighting for Long-Tailed Recognition}
\titlerunning{Optimal Shared Depth and Task Weighting for Long-Tailed Recognition}

\author{Chang Chu\inst{1}$^{\star}$ \and
Qingyue Zhang\inst{1}$^{\star}$ \and
Shao-Lun Huang\inst{1}$^{\star\star}$ \and
Junxiong Zheng\inst{2}
}
\authorrunning{C. Chu et al.}
\institute{Shenzhen International Graduate School, Tsinghua University, Shenzhen, China \and
Shenzhen Zkosemi Semiconductor Technology Co., Ltd, Shenzhen, China}

\begingroup
\renewcommand{\thefootnote}{\fnsymbol{footnote}}
\footnotetext[1]{Equal contribution.}
\footnotetext[2]{Corresponding author: Shao-Lun Huang <\email{twn2gold@gmail.com}>.}
\endgroup


\maketitle

\begin{abstract}
Long-tailed recognition suffers from a persistent head--tail trade-off: improving tail performance often degrades head accuracy and can increase training instability. Despite strong empirical results from re-weighting, decoupled training, and multi-expert methods, key design choices about representation sharing between head and tail classes and supervision weighting across class groups remain largely heuristic.
In this work, we propose OSDTW, a principled task-decomposition framework that partitions the original single-label recognition problem into a head task and a tail task, implemented with a shared encoder and task-specific decoders.
To handle the mutual exclusivity and statistical dependence between the two label groups, we introduce a factorized model and show that the resulting Kullback--Leibler divergence-based generalization error can be written as the sum of task-wise terms up to an additive constant, yielding a well-defined task-wise objective.
We further develop a three-stage training pipeline: independent task training to estimate task-wise optima and the Fisher information matrix, weighted joint training to learn a shared encoder, and branch assembly to construct the final decoupled model.
Under a block-diagonal Fisher approximation, we derive a computable second-order expansion of the expected generalization error, decomposing it into encoder variance, encoder bias, and decoder variance. This bias--variance decomposition provides a computable proxy to select the shared depth and task weights, enabling efficient hyper-parameter search. Experiments on standard long-tailed benchmarks demonstrate the effectiveness of the proposed approach over strong baselines.


\end{abstract}


\section{Introduction}

Real-world data often exhibit long-tailed label distributions: a small number of head classes have abundant samples, while most tail classes are severely under-represented \cite{liu2019large,zhou2017places,van2018inaturalist,zhang2023deep}. Under such extreme imbalance, standard empirical risk minimization is dominated by head classes, leading to insufficient tail representations, biased decision boundaries, and inter-class confusion, and ultimately a persistent head--tail trade-off. Since evaluation in practice often emphasizes balanced accuracy, long-tailed learning faces stricter performance requirements.

Existing approaches mainly include re-weighting and re-sampling, decoupled training, and multi-branch and multi-expert methods \cite{lin2017focal,cui2019class,cao2019learning,menon2020long,kang2019decoupling,zhou2020bbn,wang2020long}. While they improve tail performance to some extent, prior work has also reported side effects: re-weighting/re-sampling brings the training objective closer to the target evaluation distribution but can overfit minority classes, degrade representation quality, or increase model variance and hurt head accuracy. These findings are largely empirical, and a unified theoretical framework that explains the relationship among training weights, representation quality, and generalization error is still missing. Meanwhile, multi-branch/multi-expert architectures are widely used, yet the depth of backbone sharing is typically chosen heuristically, without theory to answer which sharing depth best benefits generalization.

\begin{figure}[h]
\centering
\subfloat[Head classes]{%
    \includegraphics[width=0.4\columnwidth]{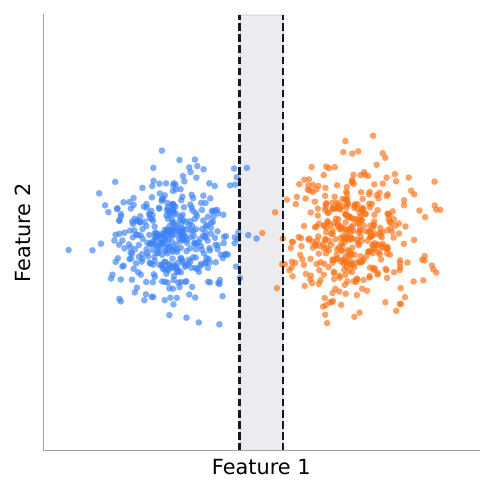}
}
\hfill
\subfloat[Tail classes]{%
    \includegraphics[width=0.4\columnwidth]{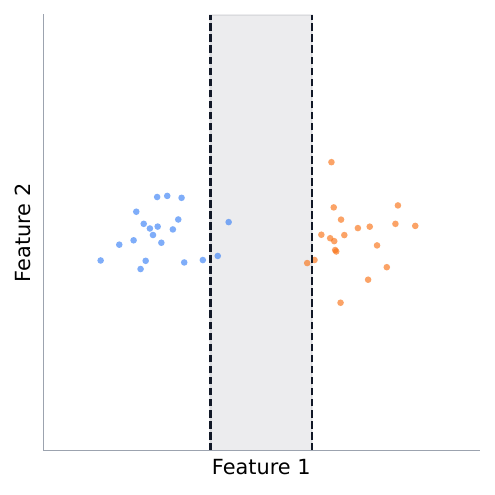}
}
\caption{Toy illustration of boundary uncertainty under finite samples. Dashed bands indicate the range of decision boundaries induced by resampling.}
\label{fig:motivation}
\end{figure}

From a more fundamental statistical view, this challenge stems from finite-sample uncertainty: head classes have more data and thus smaller estimation variance, while tail classes have fewer samples and much larger variance, as illustrated conceptually in Fig.~\ref{fig:motivation}. As a result, representation sharing can reduce variance but may introduce bias; how to make a computable trade-off between sharing and specialization, and between weighting and stability, becomes the key question.

To address these gaps, we take two steps. First, we adopt logit adjustment to handle train/test prior shift \cite{menon2020long}, and formulate the objective as minimizing K-L divergence-based generalization error \cite{zhang2026high,zhang2025asymptotic,zhang2026unified,zhang2026lora-da}. Second, we characterize head/tail task statistics via the Fisher information matrix and, under a block-diagonal approximation, establish a theoretical link among task weights, sharing depth, and generalization error, turning structural choices into a computable proxy criterion.

Based on these motivations, we propose OSDTW, a \emph{task decomposition and three-stage training} framework: we split the single-label long-tailed problem into head and tail tasks with a shared encoder and task-specific decoders; for the mutually exclusive yet dependent labels, we introduce a factorized model and show that the joint K-L divergence risk decomposes into the sum of task-wise K-L divergence terms plus a constant, yielding a well-defined task-wise objective; under the block-diagonal Fisher approximation, we derive a second-order expansion of the generalization error, explicitly decomposing it into encoder variance, encoder bias, and decoder variance, and use this expansion to select sharing depth and task weights.

\begin{itemize}
    \item To address long-tailed recognition and the head--tail trade-off, we propose OSDTW: a task-decomposition framework with a factorized model and a three-stage pipeline, which yields a task-wise K-L divergence generalization objective and a computable bias--variance proxy for selecting the shared depth and task weights.
    \item Experiments on long-tailed benchmarks demonstrate the effectiveness of OSDTW, surpassing strong baselines by 1.1\% on ImageNet-LT and 1.3\% on Places-LT.
\end{itemize}
Code and pretrained models will be released upon acceptance.

\section{Related Work}

\subsection{Re-balancing losses and sampling}
A common strategy is to re-balance training signals via re-weighting or re-sampling to mitigate head dominance under class imbalance. Representative methods include Focal loss, class-balanced weighting, logit adjustment, and LDAM~\cite{lin2017focal,cui2019class,menon2020long,cao2019learning}. These approaches can improve tail accuracy but often require heuristic hyperparameters and may increase variance on scarce classes.

\subsection{Decoupled training and classifier calibration}
Decoupled pipelines first learn representations on the imbalanced data and then retrain or calibrate the classifier with balanced sampling or post-hoc scaling~\cite{kang2019decoupling}. This remains a strong baseline for long-tailed recognition, but does not directly address how much representation should be shared across head and tail groups.

\subsection{Multi-branch and multi-expert models}
Multi-branch/multi-expert models use specialized branches or experts to trade head bias for tail variance~\cite{zhou2020bbn,wang2020long,zhang2021test}.
Examples include BBN and expert ensembles such as RIDE/TADE. Their branch depth and fusion/weighting choices are typically selected heuristically.

\subsection{Position of this work}
We cast head and tail as two explicit tasks with a shared encoder and task-specific decoders, and derive a computable bias--variance proxy that links shared depth and task weights to K-L divergence generalization. This provides a principled criterion for representation sharing and supervision balancing beyond heuristic design.

\section{Preliminaries}

\subsection{K-L divergence measure}
We adopt K-L divergence \cite{cover1999elements} as the generalization error measure.

Let $P$ and $Q$ be two distributions defined on the same probability space. The K-L divergence is defined as
\begin{align}
    D\!\left(P \middle\| Q\right)
    = \sum_{x \in \mathcal{X}} p(x)\,\log\frac{p(x)}{q(x)}.
\end{align}
In this work, we use the expectation of K-L divergence between the true distribution and the distribution learned from training samples as the generalization error, i.e.,
\begin{align}
\label{eq:kl_measure}
\mathbb{E}\!\left[D\!\left(Q_X \middle\| P_{X;\hat{\theta}}\right)\right].
\end{align}
Compared with other measures, the K-L divergence has a more direct correspondence to the cross-entropy objective \cite{zhang2025asymptotic}.

\subsection{Maximum likelihood estimation}
Consider a parametric model family $\{P_{X;\theta}:\theta\in\Theta\}$ over random variable $X$ and an independent and identically distributed (i.i.d).\ dataset $\mathcal{D}=\{x_i\}_{i=1}^n$. The MLE is defined as
\begin{align}
\label{mle_target}
\hat{\theta}_{\mathrm{MLE}}
    &= \arg\max_{{\theta} \in \Theta} \; \frac{1}{n}
    \sum_{x \in \mathcal{D}} \log P_{X;{\theta}}(x).
\end{align}
Under standard regularity conditions, the MLE is consistent and asymptotically efficient.

\subsection{Asymptotic normality}
Let $\theta^*$ denote the population maximizer of the expected log-likelihood, i.e., $\theta^*=\arg\max_{\theta}\; \mathbb{E}[\log P_{X;\theta}(X)]$. Then the MLE satisfies the classical asymptotic normality~\cite{van2000asymptotic}:
\begin{align}
\label{eq:asymptotic_normality}
\sqrt{n}\left( \hat{\theta}_{\mathrm{MLE}} - \theta^* \right)
    \xrightarrow{d} \mathcal{N}\!\left( 0, J(\theta^*)^{-1} \right),
\end{align}
where $J(\theta^*)$ is the Fisher information matrix evaluated at $\theta^*$, and the superscript $-1$ denotes the inverse matrix.

\subsection{Fisher information}
The Fisher information matrix~\cite{cover1999elements} is defined as
\begin{align}
    J({\theta})^{d \times d}
    = \mathbb{E} \left[
        \left( \frac{\partial}{\partial {\theta}} \log P_{X;{\theta}}(X) \right)
        \left( \frac{\partial}{\partial {\theta}} \log P_{X;{\theta}}(X) \right)^{\!\top}
    \right],
\end{align}
assuming $\theta\in\mathbb{R}^d$ and the score function is square-integrable.
In the well-specified setting, $J(\theta^*)$ also equals the negative Hessian of the expected log-likelihood at $\theta^*$, and $J(\theta^*)^{-1}$ characterizes the asymptotic covariance of MLE in~\eqref{eq:asymptotic_normality}.

\subsection{Mutual information and conditional mutual information}
To quantify statistical dependence, we first give the general definitions. For jointly distributed random variables $(U,V)$ with joint mass $P(u,v)$ and marginals $P(u),P(v)$, the mutual information is
\begin{align}
I(U;V)
=
\sum_{u,v} P(u,v)\,\log\frac{P(u,v)}{P(u)\,P(v)}.
\end{align}
It is non-negative and equals zero iff $U$ and $V$ are independent.
For a third variable $Y$, the conditional mutual information is
\begin{align}
I(U;V\mid Y)
=
\sum_{y,u,v} P(y,u,v)\,\log\frac{P(u,v\mid y)}{P(u\mid y)\,P(v\mid y)}.
\end{align}
In our setting, we use conditional mutual information to capture the dependence between head and tail labels under the single-label setting.

\section{Problem formulation}

We consider a $K$-class single-label recognition problem under a long-tailed class-frequency distribution.
We sort all classes by their training sample counts and partition the ordered list into two disjoint groups with the same number of classes: the more frequent half forms Task~A (head), and the remaining half forms Task~B (tail). When $K$ is odd, the two groups differ by one class.
Let $Y$ denote the input and let $Z\in\{0,1\}^K$ be the one-hot label. We split $Z$ into two parts,
$Z_A$ and $Z_B$, corresponding to head and tail classes, respectively. The two label vectors are
mutually exclusive and statistically dependent given $Y$.
We model each task with a Bernoulli factorization (BCE loss), so an all-zero $Z_t$ is naturally supported and indicates that the sample belongs to the other task's label group.
Define
\begin{align}
X_A=(Y,Z_A),\quad X_B=(Y,Z_B),\quad W=(Y,Z_A,Z_B),
\end{align}
and assume the training data $W^N=\{w_i\}_{i=1}^N$ are i.i.d.\ samples from an underlying distribution $Q_W$.
The task-specific projections are
\begin{align}
X_A^N=\{x_{A,i}\}_{i=1}^N,\qquad X_B^N=\{x_{B,i}\}_{i=1}^N.
\end{align}
We denote the marginal of $Q_W$ by $Q_Y$. All parameters (e.g., $\theta,\phi,\psi$) are vectors unless otherwise stated, and weights $w_A,w_B$ are scalars.

Our goal is to minimize the K-L divergence generalization error.
To make the objective explicit for head/tail tasks, we introduce a task decomposition that yields
a tractable risk decomposition.

\textbf{Factorized model and decoupled parameterization.}
We introduce a factorized model
\begin{align}
P_{Z_A,Z_B\mid Y}(z_A,z_B\mid y)
=
P_{Z_A\mid Y;\theta^A}(z_A\mid y)\,P_{Z_B\mid Y;\theta^B}(z_B\mid y),
\end{align}
and define the induced marginal model
\begin{align}
P_{X_t;\theta^t}(y,z_t)=Q_Y(y)\,P_{Z_t\mid Y;\theta^t}(z_t\mid y),\quad t\in\{A,B\}.
\end{align}
Each task uses a decoupled parameterization $\theta^t=(\phi^t,\psi^t)$ with a shared encoder depth $C$.

\textbf{Task-wise K-L divergence via a decomposition.}
The factorization does not ignore dependency; instead, the dependency contributes only an additive constant, as stated below:
\begin{lemma}[Joint K-L divergence decomposition]
\label{lem:kl_mi}
For any true conditional joint $Q_{Z_A,Z_B\mid Y}$ and any factorized predictor $P_AP_B$,
\begin{align}
\KL(Q_W\Vert P_W)
= {} \KL(Q_{X_A}\Vert P_{X_A})
+ \KL(Q_{X_B}\Vert P_{X_B})
+ I_Q(Z_A;Z_B\mid Y),
\end{align}
where $I_Q(Z_A;Z_B\mid Y)\ge 0$ is a constant independent of $(C,w_A,w_B)$.
\end{lemma}
Hence, minimizing the joint K-L divergence risk is equivalent to minimizing the \emph{task-wise} generalization error
\begin{align}
\mathcal{R}
\triangleq
\mathbb{E}\!\left[\KL\!\left(Q_{X_A}\,\|\,P_{X_A;\hat{\theta}^A}\right)\right]
+
\mathbb{E}\!\left[\KL\!\left(Q_{X_B}\,\|\,P_{X_B;\hat{\theta}^B}\right)\right].
\end{align}

\section{Main Result}

\begin{figure*}[htbp]
\centering
\includegraphics[width=\textwidth]{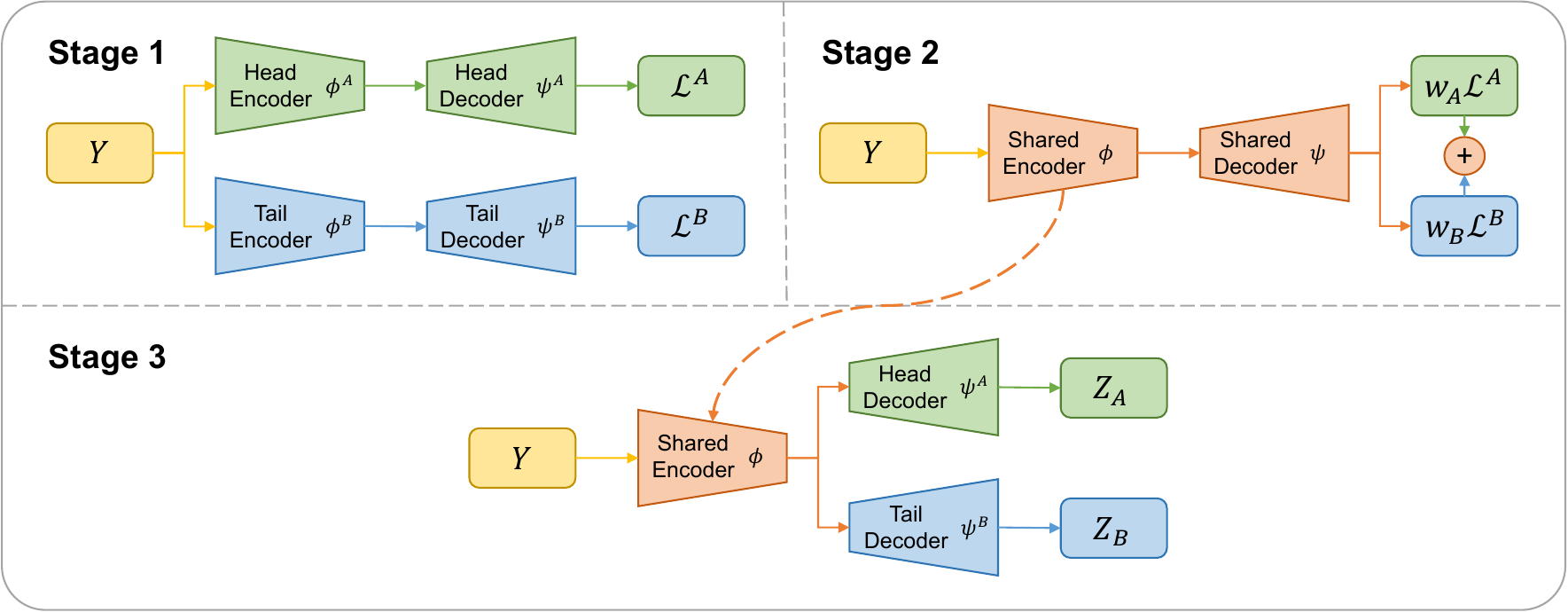}
\caption{Three-stage training with a shared encoder of depth $C$ and task-specific decoders.}
\label{arch}
\end{figure*}

\textbf{Three-stage training pipeline.}
We adopt a three-stage pipeline, as illustrated in Fig.~\ref{arch}, to decouple statistic estimation,
structure selection, and final assembly, making the shared depth $C$ and task weights $w_A,w_B$ explicit.
The overall procedure is summarized in Algorithm~\ref{alg:osdtw_training}.

Stage~1 trains each task independently:
\begin{align}
(\hat\phi^t,\hat\psi^t)=\arg\max_{\phi,\psi}\ \sum_{x\in X_t^N}\log P_{X_t;(\phi,\psi)}(x),
\qquad t\in\{A,B\}.
\end{align}
This stage provides task-level statistics such as Fisher blocks, which are then used to select the shared depth $C$ and the task weights $w_A,w_B$.

Stage~2 performs weighted training on the full network to obtain an optimal shared representation, using the objective
\begin{align}
(\hat\phi,\hat\psi)
&=
\arg\max_{\phi,\psi}
\Big[
 w_A\sum_{x\in X_A^N}\log P_{X_A;(\phi,\psi)}(x)
\nonumber
+ w_B\sum_{x\in X_B^N}\log P_{X_B;(\phi,\psi)}(x)
\Big].
\label{eq:stage2_objective}
\end{align}
Here $w_A,w_B$ balance head/tail supervision and satisfy $w_A+w_B=1$. After training, we truncate the first $C$ layers of the
resulting network as the shared encoder.

Stage~3 assembles the final two-branch predictor by combining the shared encoder learned in Stage~2 with the task-specific decoders learned in Stage~1:
\begin{align}
\hat{\theta}^A=(\hat{\phi},\hat{\psi}^A),\qquad
\hat{\theta}^B=(\hat{\phi},\hat{\psi}^B).
\end{align}
This preserves low-variance shared features while keeping task-specific decoders for bias reduction.

\begin{algorithm}[t]
\caption{Three-Stage Training, Structure Selection, and Inference of OSDTW}
\label{alg:osdtw_training}
\footnotesize
\begin{algorithmic}[1]
\REQUIRE Task datasets $X_A^N,X_B^N$, candidate shared depths $\mathcal{C}$, candidate task weights $\mathcal{W}$, optional test sample $y$
\ENSURE Final two-branch model $(\hat{\theta}^A,\hat{\theta}^B)$ and predicted class $\hat{k}$ for $y$ if provided
\STATE \textbf{Stage 1: Independent task training}
\FOR{$t\in\{A,B\}$}
    \STATE Train task $t$ independently and obtain $(\hat{\phi}^t,\hat{\psi}^t)$
    \STATE Estimate task-wise statistics required by the proxy, including $J_{\phi\phi}^t$ and $\phi^{t*}$
\ENDFOR
\STATE Approximate the encoder mismatch by $\Delta=\phi^{B*}-\phi^{A*}$
\STATE \textbf{Proxy-based structure selection}
\FOR{each shared depth $C\in\mathcal{C}$}
    \FOR{each task weight $w_A\in\mathcal{W}$ with $w_B=1-w_A$}
        \STATE Evaluate the theoretical proxy of $\mathcal{R}$ using the estimated statistics
    \ENDFOR
\ENDFOR
\STATE Select $(C^\star,w_A^\star,w_B^\star)$ with the smallest proxy value
\STATE \textbf{Stage 2: Weighted joint training}
\STATE Train the full network with weights $(w_A^\star,w_B^\star)$ to obtain $(\hat{\phi},\hat{\psi})$
\STATE Extract the first $C^\star$ encoder layers as the shared encoder
\STATE \textbf{Stage 3: Branch assembly}
\STATE Form the final predictor as $\hat{\theta}^A=(\hat{\phi},\hat{\psi}^A)$ and $\hat{\theta}^B=(\hat{\phi},\hat{\psi}^B)$
\STATE \textbf{Optional practical refinement}
\STATE Fine-tune only the task-specific decoders while freezing the shared encoder
\IF{a test sample $y$ is given}
    \STATE Compute the shared representation $h=\hat{\phi}(y)$
    \STATE Compute task-wise logits $s^A=\hat{\psi}^A(h)$ and $s^B=\hat{\psi}^B(h)$
    \STATE Assemble the final logit vector $s=[s^A;s^B]$
    \STATE Predict the class by $\hat{k}=\arg\max_k s_k$
\ENDIF
\RETURN $(\hat{\theta}^A,\hat{\theta}^B)$ and $\hat{k}$ if available
\end{algorithmic}
\end{algorithm}

We present a computable bias--variance characterization for representation sharing in long-tailed single-label recognition. Our key claim is that, under a principled task decomposition and a factorized model, the expected K-L divergence generalization gap of the final assembled two-branch model admits an explicit second-order expansion. This yields a direct proxy to select the shared depth $C$ and the task weights $(w_A,w_B)$.

\textbf{Block Fisher and local mismatch.}
Let $J^t$ be the task-wise Fisher information at the population optimum. Following common scalable second-order approximations that neglect cross-block couplings to obtain tractable curvature surrogates~\cite{martens2015optimizing}, we adopt a block-diagonal Fisher approximation:
\begin{align}
J^t=
\begin{bmatrix}
J_{\phi\phi}^t & 0\\
0 & J_{\psi\psi}^t
\end{bmatrix},\qquad t\in\{A,B\},
\end{align}
and denote $d_\psi^t=\dim(\psi^t)$.
Let $\phi^{t*}$ denote the population-optimal encoder for task $t\in\{A,B\}$, and define
$\Delta\triangleq \phi^{B*}-\phi^{A*}$ as the encoder mismatch. To facilitate the second-order asymptotic analysis below, we adopt a local mismatch assumption
$\|\Delta\|=O(N^{-1/2})$. This assumption is made under the premise that the head and tail tasks are both derived from the same long-tailed classification problem and therefore share a substantial amount of representation structure. Empirically, prior studies have shown that early and intermediate layers tend to capture transferable features across related tasks, and keeping parameters close to a reference solution can improve fine-tuning stability and generalization~\cite{raghu2019transfusion,lee2020mixout,chen2020recall}. Similar local-neighborhood assumptions have also been adopted in related asymptotic analyses and transfer-learning theory~\cite{zhang2026high,zhang2025asymptotic,zhang2026unified,zhang2026lora-da}.

\begin{theorem}[Expected K-L divergence generalization error under block-diagonal Fisher and local mismatch]
\label{thm:main_result}
Under the block-diagonal Fisher approximation and the local mismatch regime above, define
\begin{align}
H(w)\triangleq w_A\,J_{\phi\phi}^A+w_B\,J_{\phi\phi}^B,
\qquad
G(w)\triangleq w_A^2\,J_{\phi\phi}^A+w_B^2\,J_{\phi\phi}^B.
\end{align}
Then the expected generalization error admits the second-order approximation
\begin{align}
\label{eq:main_result_expansion}
\begin{aligned}
&
\underbrace{\frac{1}{2N}\tr\!\Big((J_{\phi\phi}^A+J_{\phi\phi}^B)\,H(w)^{-1}G(w)H(w)^{-1}\Big)}_{\text{encoder variance}}
\\
&+
\underbrace{\frac{1}{2}\Delta^\top\Big(w_B^2\, J_{\phi\phi}^A+w_A^2\,J_{\phi\phi}^B\Big)\Delta}_{\text{encoder bias}}
+
\underbrace{\frac{d_\psi^A+d_\psi^B}{2N}}_{\text{decoder variance}}
+
o\!\left(\frac{1}{N}\right).
\end{aligned}
\end{align}
\end{theorem}

\textbf{Design guideline.}
Eq.~\eqref{eq:main_result_expansion} provides a practical selection rule for $(C,w_A)$.
Increasing the shared depth $C$ typically reduces the \emph{encoder variance} term via stronger joint estimation, but may enlarge the \emph{encoder bias} term when the task-optimal encoders differ (larger $\Delta$). The weight $(w_A,w_B)$ controls the variance--bias balance between head and tail supervision. In practice, Stage~1 supplies empirical estimates of $J_{\phi\phi}^A,J_{\phi\phi}^B$ and $\Delta$, enabling efficient search over $(C,w_A)$ by minimizing the proxy in~\eqref{eq:main_result_expansion}.

\section{Experiments}

\begin{table}[htbp]
\centering
\caption{Top-1 accuracy (\%) on ImageNet-LT.}
\renewcommand\arraystretch{0.95}
\begin{tabular}{p{3cm}cccc}
\toprule
Method & Many & Medium & Few & All\\
\midrule
\rowcolor{Gray} \multicolumn{5}{c}{ResNet50 Backbone} \\
\midrule
CB \cite{cui2019class} & 64.0 & 33.8 & 5.8 & 41.6\\
LDAM \cite{cao2019learning} & 60.4 & 46.9 & 30.7 & 49.8\\
Logit Adj. \cite{menon2020long} & 61.1 & 47.5 & 27.6 & 50.1\\
RIDE(4E) \cite{wang2020long} & 68.3 & 53.5 & 35.9 & 56.8\\
MiSLAS \cite{zhong2021improving} & 62.9 & 50.7 & 34.3 & 52.7\\
DisAlign \cite{zhang2021distribution} & 61.3 & 52.2 & 31.4 & 52.9\\
PaCo \cite{cui2021parametric} & 68.0 & 56.4 & 37.2 & 58.2\\
GCL \cite{li2022long} & 63.0 & 52.7 & 37.1 & 54.5\\
\midrule
\rowcolor{Gray} \multicolumn{5}{c}{ViT-B Backbone} \\
\midrule
MAE \cite{he2022masked} & 74.7 & 48.2 & 19.4 & 54.5\\
DeiT \cite{touvron2022deit} & 70.4 & 40.9 & 12.8 & 48.4\\
DeiT-LT \cite{rangwani2024deit} & 66.6 & \textbf{58.3} & 40.0 & 59.1\\
PaCo \cite{cui2021parametric} & 70.4 & 57.0 & \textbf{43.5} & 60.3\\
LiVT \cite{xu2023learning} & 75.8 & 56.2 & 32.1 & 60.9\\
OSDTW & \textbf{76.3} & \textbf{58.3} & 34.9 & \textbf{62.0}\\
\bottomrule
\end{tabular}
\label{tab:performance-imagenet-lt}
\end{table}

\begin{table}[htbp]
\centering
\caption{Top-1 accuracy (\%) on Places-LT.}
\renewcommand\arraystretch{0.95}
\begin{tabular}{p{3cm}cccc}
\toprule
Method & Many & Medium & Few & All\\
\midrule
\rowcolor{Gray} \multicolumn{5}{c}{ResNet152 Backbone} \\
\midrule
CB \cite{cui2019class} & 45.7 & 27.3 & 8.2 & 30.2\\
Focal \cite{lin2017focal} & 41.1 & 34.8 & 22.4 & 34.6\\
DisAlign \cite{zhang2021distribution} & 40.4 & 42.4 & 30.1 & 39.3\\
LADE \cite{hong2021disentangling} & 42.8 & 39.0 & 31.2 & 38.8\\
RSG \cite{wang2021rsg} & 41.9 & 41.4 & 32.0 & 39.3\\
TADE \cite{zhang2021test} & 43.1 & 42.4 & 33.2 & 40.9\\
PaCo \cite{cui2021parametric} & 36.1 & \textbf{47.9} & \textbf{35.3} & 41.2\\
GCL \cite{li2022long} & - & - & - & 40.6\\
\midrule
\rowcolor{Gray} \multicolumn{5}{c}{ViT-B Backbone} \\
\midrule
MAE \cite{he2022masked} & 48.9 & 24.6 & 8.7 & 30.3\\
DeiT \cite{touvron2022deit} & \textbf{51.6} & 31.0 & 9.4 & 34.2\\
PaCo \cite{cui2021parametric} & \textbf{51.6} & 32.6 & 13.2 & 35.5\\
LiVT \cite{xu2023learning} & 48.1 & 40.6 & 27.5 & 40.8\\
OSDTW & 49.7 & 41.8 & 30.9 & \textbf{42.5}\\
\bottomrule
\end{tabular}
\label{tab:performance-place}
\end{table}

\subsection{Datasets}
We evaluate the proposed method on two long-tailed benchmarks, ImageNet-LT and Places-LT, and conduct ablation studies on CIFAR-100-IR100.
ImageNet-LT and Places-LT are long-tailed subsets derived from ImageNet and Places, respectively~\cite{liu2019large,zhou2017places}.
CIFAR-100-IR100 is constructed from CIFAR-100 by applying an imbalance ratio of 100~\cite{krizhevsky2009learning}.

\subsection{Implementation details}
We use ViT-Base (12 layers) as the backbone \cite{dosovitskiy2020image}. Stage~1 and Stage~2 each train for
100 epochs. After Stage~3, we further fine-tune only the task-specific decoders for 30 epochs while freezing the shared encoder. This is a lightweight post-assembly refinement used only in implementation, and does not affect the proxy-based selection of $C$ and $w_A,w_B$.
All experiments are conducted on 4 NVIDIA GeForce RTX 4090 GPUs.
In Stage~1, we estimate the Fisher information matrix using the empirical diagonal Fisher, i.e., the average squared score gradients over training samples. This diagonal empirical Fisher provides a tractable measure of parameter sensitivity and has been used to improve low-rank compression~\cite{hsu2022language}, quantization~\cite{kim2023squeezellm}, and fine-tuning~\cite{guo2023lq} of pretrained language models, which suffices for capturing the relative parameter sensitivity of the two tasks in our framework.

For a fair comparison, we follow the common setup used by LiVT~\cite{xu2023learning} and GPaCo/PaCo~\cite{cui2023generalized,cui2021parametric}, using the same MAE-pretrained model~\cite{he2022masked} trained for 800 epochs with a mask ratio of 0.75, and keep the same hyperparameters: optimizer AdamW(0.9, 0.99), learning rate 1e-3, batch size 1024,
input size 224, drop path 0.1, weight decay 0.5, and data augmentation RandAug(9, 0.5), Mixup(0.8),
and CutMix(1.0). We re-evaluate and report the results of LiVT and the strong ResNet-based baseline PaCo on the ViT-B backbone, while the results of other baselines are taken from their original papers.
During training, we apply logit adjustment using training-set class priors to address prior shift.

\subsection{Comparison to prior work}

Table~\ref{tab:performance-imagenet-lt} shows the experimental result on ImageNet-LT. Our method reaches 62.0\% top-1 accuracy, surpassing the strongest baseline LiVT by 1.1\%.
Table~\ref{tab:performance-place} shows the experimental result on Places-LT. Our method reaches 42.5\% top-1 accuracy, surpassing the strongest available baseline PaCo by 1.3\%.
We also observe that LiVT and OSDTW achieve better performance than PaCo on the ViT-B backbone, which is consistent with the findings reported by LiVT that BCE-based methods can outperform CE-based methods for long-tailed recognition under ViT-B.

\subsection{Ablation study}

\begin{figure}[htbp]
\centering
\includegraphics[width=\textwidth]{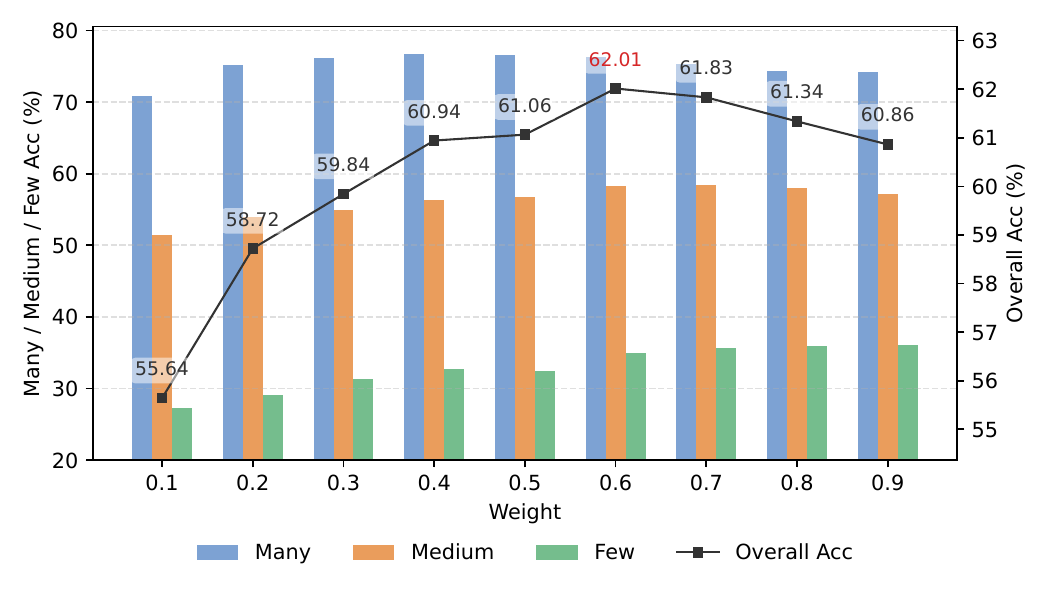}
\caption{ImageNet-LT accuracy of ViT-B under different task weights with full sharing ($C=12$).}
\label{fig:imagenet_weight}
\end{figure}

On ImageNet-LT with full sharing ($C=12$), Fig.~\ref{fig:imagenet_weight} shows that the overall accuracy first
increases and then decreases as $w_A$ grows, peaking at 62.0\% when $w_A=0.6$. We also observe a clear asymmetry:
assigning more weight to head classes consistently outperforms assigning more weight to tail classes, which contrasts
with common long-tailed reweighting heuristics. We attribute this to the limited information in tail classes:
overweighting tail supervision inflates variance and hurts overall generalization. Interestingly, the few-shot accuracy
increases monotonically with $w_A$, suggesting that stronger head supervision can improve tail generalization.

\begin{figure}[htbp]
\centering
\subfloat[Top-1 accuracy]{%
    \includegraphics[width=0.48\columnwidth]{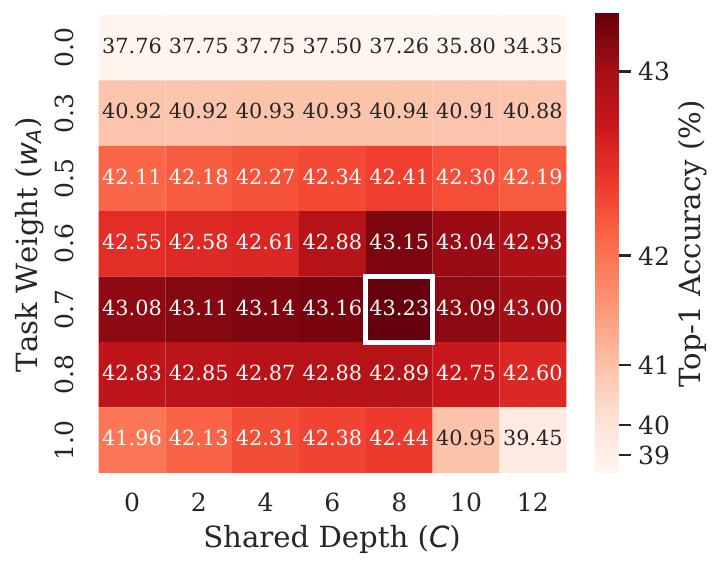}
}
\hfill
\subfloat[Generalization error]{%
    \includegraphics[width=0.48\columnwidth]{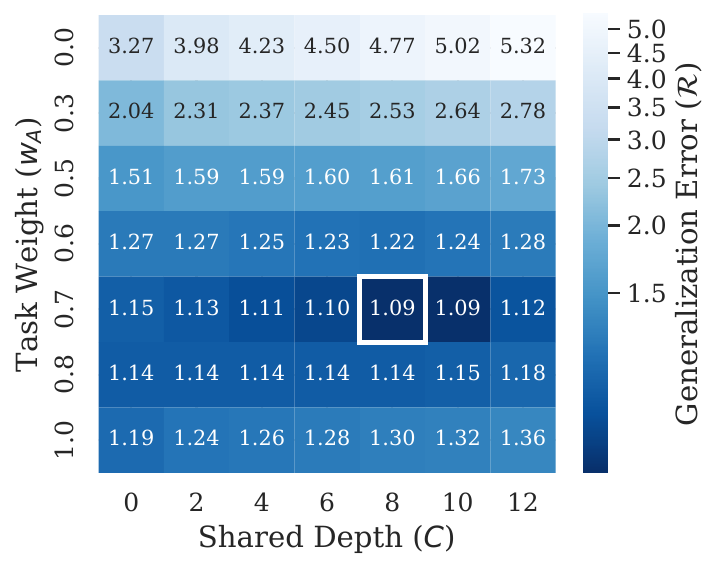}
}
\caption{CIFAR-100-IR100 accuracy and generalization error of ViT-B across shared depth $C$ and task weight $w_A$, with the optimal combination highlighted by a white box.}
\label{fig:grid_search_heatmap}
\end{figure}

On CIFAR-100-IR100, Fig.~\ref{fig:grid_search_heatmap} reports the joint effect of shared depth $C$ and task weight
$w_A$ on accuracy and generalization error. Both metrics are highly consistent and achieve the best performance
at $C=8, w_A=0.7$, which is about 1\% better than $C=12, w_A=0.5$. This supports the validity of our generalization
proxy and the need to jointly optimize depth and weight. Moreover, task weights have a stronger impact than shared
depth: optimizing $C$ alone yields about a 0.2\% gain, while optimizing $w_A$ alone yields about a 0.8\% gain.

\subsection{Runtime analysis}
Table~\ref{tab:runtime} reports the runtime of each component in our pipeline on CIFAR-100-IR100, ImageNet-LT, and Places-LT.
The proxy-based search over $C\in\{0,1,\ldots,12\}$ and $w_A\in\{0,0.1,\ldots,1.0\}$ takes only 1.18\,s on
CIFAR-100-IR100, 1.22\,s on ImageNet-LT, and 1.14\,s on Places-LT, which is negligible compared with Stage~1 and Stage~2 training.
The total runtime is 378.34\,s on CIFAR-100-IR100, 17195.94\,s on ImageNet-LT, and 2526.20\,s on Places-LT.

\begin{table}[t]
\centering
\caption{Runtime analysis of ViT-B on CIFAR-100-IR100, ImageNet-LT, and Places-LT (seconds).}
\footnotesize
\renewcommand\arraystretch{0.9}
\begin{tabular*}{0.8\textwidth}{@{\extracolsep{\fill}}lccccc}
\toprule
Dataset & Stage~1 & Search & Stage~2 & Refinement & Total\\
\midrule
CIFAR-100-IR100 & 178.51 & 1.18 & 160.03 & 38.62 & 378.34\\
ImageNet-LT & 7874.79 & 1.22 & 7157.64 & 2162.29 & 17195.94\\
Places-LT & 1193.07 & 1.14 & 1015.74 & 316.25 & 2526.20\\
\bottomrule
\end{tabular*}
\label{tab:runtime}
\end{table}

\section{Conclusion}
We proposed OSDTW, a task-decomposition framework for long-tailed recognition that yields a well-defined task-wise K-L divergence objective via a factorized model and a three-stage training pipeline. Under a block-diagonal Fisher approximation, we derived a computable bias--variance proxy to select shared depth and task weights, and demonstrated consistent gains on standard long-tailed benchmarks.

\bibliographystyle{splncs04}
\bibliography{ref}

\end{document}